\documentclass[letterpaper, 10 pt, conference]{ieeeconf}  

\IEEEoverridecommandlockouts                              

\usepackage{graphics} 
\usepackage{epsfig} 
\usepackage{times} 
\usepackage{amsmath} 
\usepackage{amssymb}  
\usepackage{adjustbox}
\usepackage{multirow}
\usepackage{hhline}
\usepackage{array}
\usepackage{makecell}
\usepackage{bibentry}
\usepackage{epstopdf,cite}
\usepackage{xcolor}
\usepackage{color}
\usepackage{bbding}
\usepackage{pifont}
\usepackage{lipsum}
\usepackage{mathtools}

\title{\LARGE \bf
RCM-Fusion: Radar-Camera Multi-Level Fusion for 3D Object Detection
}

\author{Jisong Kim$^{1, \dagger}$, Minjae Seong$^{2, \dagger}$, Geonho Bang$^{2}$, Dongsuk Kum$^{3}$ and Jun Won Choi$^{1, \ast}$
\thanks{$^{1}$Department of Electrical Engineering, Hanyang University, Seoul, 04763, South Korea. {\footnotesize jskim@spa.hanyang.ac.kr}, {\footnotesize junwchoi@hanyang.ac.kr}}
\thanks{$^{2}$Department of Artificial Intelligence, Hanyang University, Seoul, 04763, South Korea. {\footnotesize \{mjseong, ghbang\}@spa.hanyang.ac.kr}}
\thanks{$^{3}$ Graduate School of Mobility, Korea Advanced Institute of Science and Technology (KAIST), Daejeon 34141, South Korea. {\footnotesize dskum@kaist.ac.kr}}
\thanks{$\dagger$ Equally contributed authors}
\thanks{$\ast$ Corresponding author}
}

\begin{document}

\maketitle
\thispagestyle{empty}
\pagestyle{empty}


\begin{abstract}
While LiDAR sensors have been successfully applied to 3D object detection, the affordability of radar and camera sensors has led to a growing interest in fusing radars and cameras for 3D object detection. However, previous radar-camera fusion models could not fully utilize the potential of radar information. In this paper, we propose {\it Radar-Camera Multi-level fusion} (RCM-Fusion), which attempts to fuse both modalities at feature and instance levels. For feature-level fusion, we propose a {\it Radar Guided BEV Encoder} which transforms camera features into precise BEV representations using the guidance of radar Bird’s-Eye-View (BEV) features and combines the radar and camera BEV features. For instance-level fusion, we propose a {\it Radar Grid Point Refinement} module that reduces localization error by accounting for the characteristics of the radar point clouds. The experiments on the public nuScenes dataset demonstrate that our proposed RCM-Fusion achieves state-of-the-art performances among single frame-based radar-camera fusion methods in the nuScenes 3D object detection benchmark. The code will be made publicly available.
\end{abstract}

\IEEEpeerreviewmaketitle
\section{INTRODUCTION}
Accurate perception of the surrounding environments is crucial for successfully implementing self-driving robots or vehicles. Over the last decade, research into 3D object detection has focused on leveraging LiDAR's precise spatial information and camera's rich semantic information. However, LiDAR sensors may render them impractical due to their high cost, so radar sensors are emerging as a low-cost alternative. Radar sensors use radio waves to detect, locate and track objects in their vicinity. Radar sensors furnish both the 3D location of an object and its attributes, including Radar Cross Section (RCS) and Doppler velocity information. The 77GHz radars used for automotive applications have a longer wavelength than the LiDAR, making them more reliable in adverse weather conditions. Utilizing the advantages of these radar sensors, the previous literature has demonstrated that radar-camera fusion models can yield performance comparable to LiDAR-based 3D object detectors.

\begin{figure}[t]
    \centering
        \includegraphics[width=0.99\linewidth]{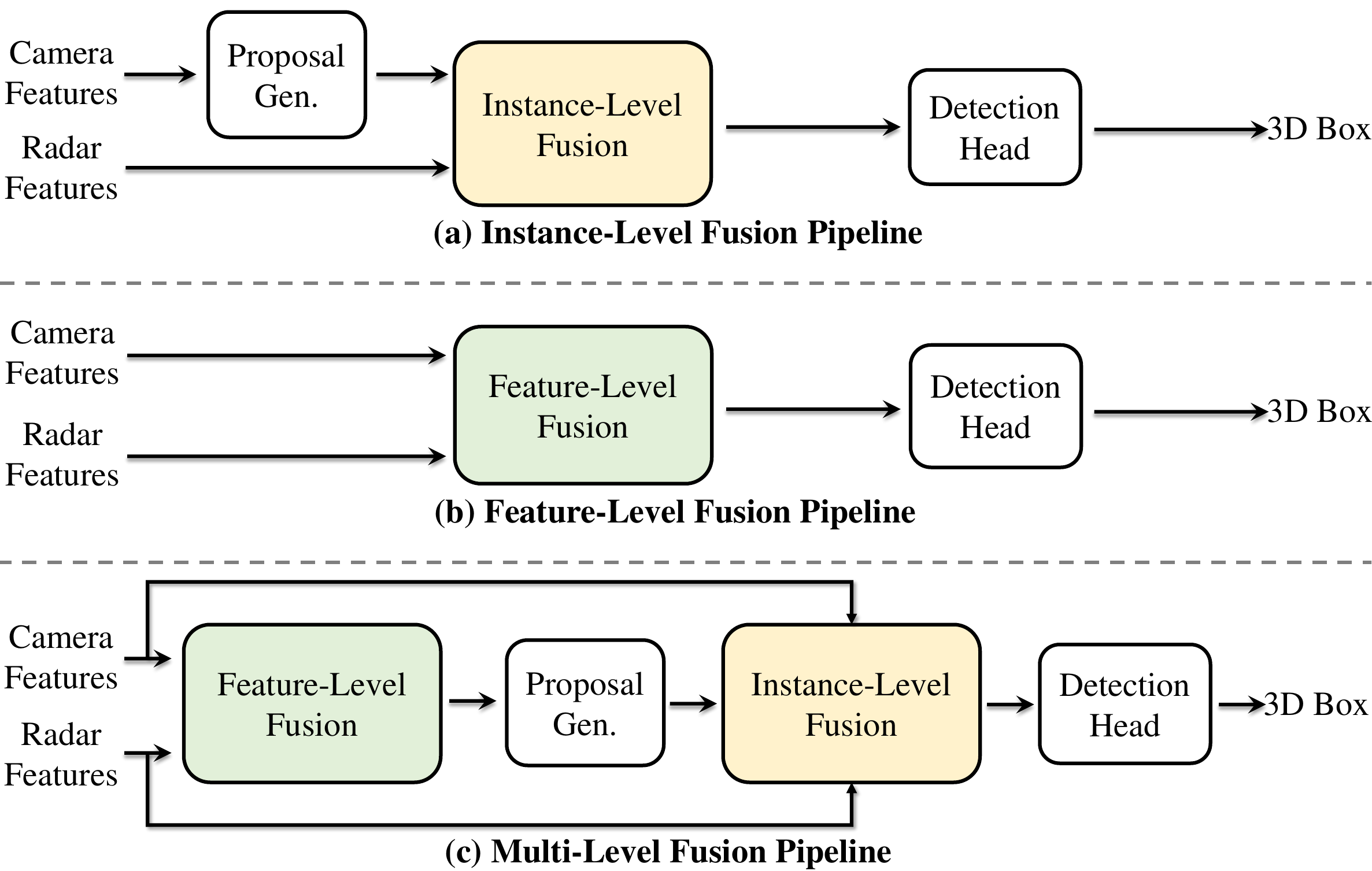}
        \caption{\textbf{Radar-camera fusion pipeline:} Previous radar-camera fusion methods have primarily focused on fusion at single levels, such as (a) or (b). These approaches inherently suffer from limitations due to this level-specific fusion approach. To address this issue, we introduce a novel multi-level fusion method (c). By integrating multimodal information at multiple levels, this novel approach effectively overcomes the limitations of previous single-level fusion methods.}
    \label{fig:intro}
\end{figure}

Recent advancements in radar-camera fusion have led to notable improvements in 3D object detection. These methods are roughly categorized into two distinct paradigms: instance-level fusion \cite{craft, RADIANT, Centerfusion} and feature-level fusion \cite{crn}. The instance-level fusion methods produce 3D box proposals from the camera features and associate the radar features represented in point clouds with 3D proposals in a camera-view to refine the initial results (see Fig.~\ref{fig:intro}(a)). Despite their strengths, instance-level methods have a notable limitation: they do not fully exploit the radar's capabilities during the initial proposal generation stage. 
Recently, feature-level fusion method, transformed the camera features into a bird's eye view (BEV) domain through Lift, Splat, and Shoot (LSS) module \cite{LSS} which was then fused with the radar BEV features (see Fig.~\ref{fig:intro}(b)).
Relying solely on either feature-level or instance-level fusion is limited in leveraging radar point information. To address this, we introduce a multi-level fusion framework that exploits radar and camera data at both the feature and instance levels (see Fig.~\ref{fig:intro}(c)).

In this paper, we propose a new radar-camera fusion architecture, referred to as {\it Radar-Camera Multi-level Fusion (RCM-Fusion)}  for 3D object detection. The proposed method achieves the fusion of radar and camera features at two different levels. RCM-Fusion first attempts feature-level fusion to produce a dense BEV feature map with which 3D proposals are generated. Then, it performs the instance-level fusion to refine the proposals through a grid point-based proposal feature fusion.

We devise Radar Guided BEV Query for feature-level fusion, which utilizes radar positional information to transform image features into distinct BEV features. Then, we employ the Radar-Camera Gating module to obtain the weighted aggregation of multi-modal BEV feature maps. Such an adaptive feature aggregation module is integrated into transformer layers, which successively decode the dense BEV query features. For instance-level fusion, we propose the Proposal-aware Radar Attention module to obtain radar point features that consider their relevance to the 3D proposal. Then, we employ a novel adaptive grid point pooling method, Radar Grid Point Pooling, which accounts for the positional uncertainty of radar points for instance-level fusion. Finally, to improve the efficacy of radar data utilization and obtain high-quality radar points for RCM-Fusion, we modified the basic radar data filtering and multi-sweep methods used in the nuScenes dataset \cite{nuscenes}. As a result, our model significantly outperformed camera-only baseline models and achieved state-of-the-art performance among the radar-camera fusion methods on the nuScenes 3D object detection benchmark \cite{nuscenes}.

The key contributions of our work are as follows:
\begin{itemize}
    \item We present a state-of-the-art radar-camera fusion method for 3D object detection. The proposed RCM-Fusion combines radar and camera features at both feature and instance levels. This is in contrast to the existing radar-camera fusion methods \cite{craft, RADIANT,Centerfusion,crn}, which perform specific-level fusion only. 
    \\
    \item We propose a {\bf feature-level fusion} which transforms the image features into accurate BEV features through Radar Guided BEV Query and adaptively fuses multi-modal BEV features via Radar-Camera Gating modules. This feature-level fusion generates a dense BEV feature map, which is used to produce the initial 3D proposals. 
    \\
    \item We propose an {\bf instance-level fusion} that employs Proposal-aware Radar Attention to reduce the impact of irrelevant points and generates features which refine the initial results through Radar Grid Point Pooling.
    \\
    \item RCM-Fusion records a state-of-the-art performance of {\bf 50.6\%} mAP and {\bf 58.7\%} NDS in single frame-based radar-camera fusion methods on the nuScenes test set.

\end{itemize}

\section{RELATED WORK}

\subsection{Radar-camera sensor fusion for 3D object detection}

Radar-camera fusion models have been conducted in order to leverage radar positional information to supplement the camera's limited depth information. The challenge of radar-camera fusion is to devise a strategy that effectively integrates data from various viewpoints. GriFNet \cite{GrifNet} utilizes 3D anchors and 3D proposals to effectively extract features from radar and camera data. CramNet \cite{CramNet}, meanwhile, leverages a 2D segmentation model to define the foreground section of both features and then fuses them in a unified 3D environment. CenterFusion \cite{Centerfusion}, CRAFT \cite{craft} reduce localization and velocity errors by employing radar information based on camera-based 3D object detection results. RADIANT \cite{RADIANT} designed a network to estimate the positional offset between the radar returns and the object center in order to address the radar-camera association problem. CRN \cite{crn} utilizes camera BEV features, which are transformed from perspective view camera features using radar assistance, and fuses them with radar BEV features for 3D object detection. 

Our RCM-Fusion differs from the aforementioned methods in that it performs radar-camera fusion at both feature and instance levels. Additionally, while the recent feature-level fusion method, CRN \cite{crn}, necessitates depth ground truth labels for training its depth prediction network, our approach does not have this requirement.

\subsection{Grid-based refinement for two-stage 3D object detection}
 Two-stage 3D object detectors \cite{voxelrcnn,  PyramidRCNN, pvrcnn, GraphRCNN} using LiDAR data have been actively developed to refine initial prediction results by leveraging LiDAR point clouds within proposal boxes. In \cite{voxelrcnn, PyramidRCNN, pvrcnn}, virtual grid points were used to refine initial results.  While PV-RCNN \cite{pvrcnn} and Voxel R-CNN \cite{voxelrcnn} used uniformly spaced virtual grid points for refinement, Pyramid R-CNN \cite{PyramidRCNN} proposed a method that can reliably refine even when the internal points in proposals are sparse and irregularly distributed. 

 Our RCM-Fusion uses a novel grid-based refinement approach, which adaptively constructs 2D virtual grids, taking into account the directional uncertainty associated with radar measurements.

\begin{figure*}[t]
    \centering
        \includegraphics[width=0.96\linewidth]{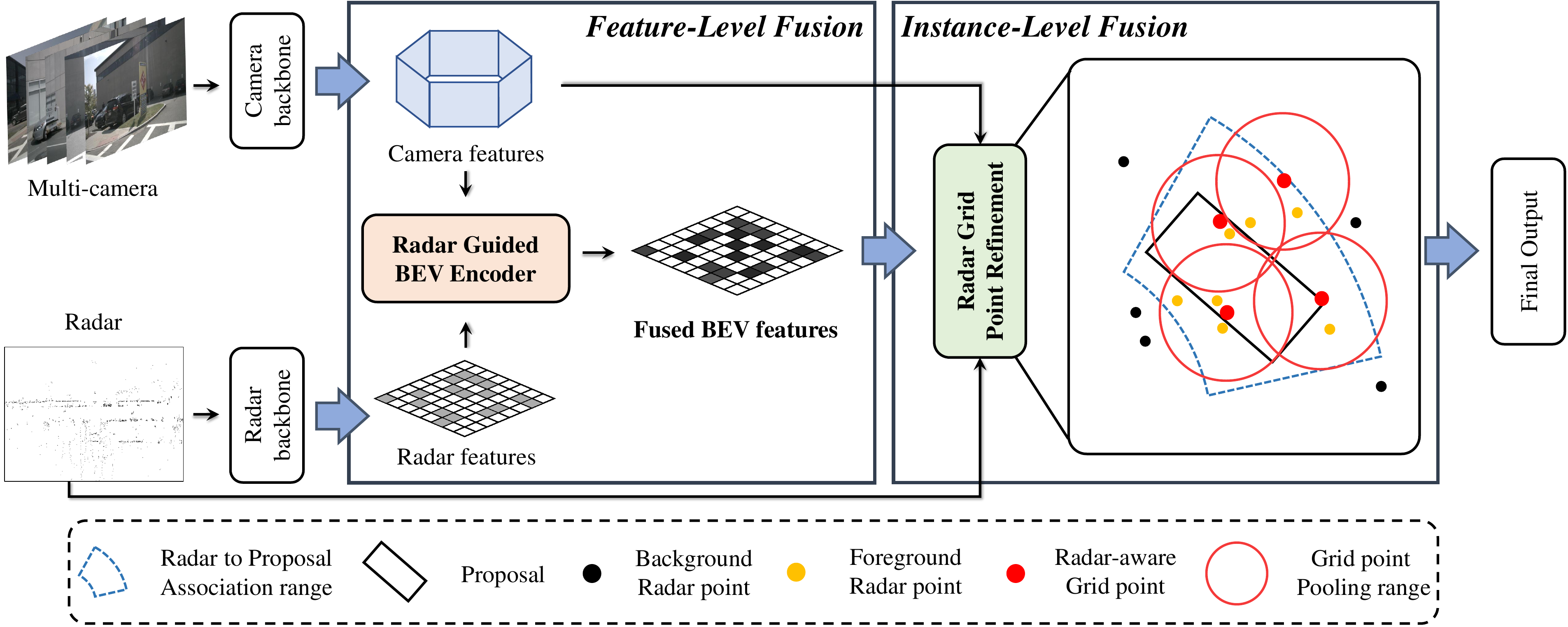}
        \caption{\textbf{Overall architecture of the proposed RCM-Fusion:} RCM-Fusion uses each backbone network to obtain the radar features and camera features. The multi-modal features are then fused in the Radar Guided BEV Encoder module for feature-level fusion. For instance-level fusion, the Radar Grid Point Refinement module refines the initial results with a novel grid feature pooling method.}
    \label{fig:overall}
\end{figure*}

\section{PROPOSED METHOD}
We propose a multi-level fusion approach to efficiently fuse radar point clouds and surrounding images. Fig.~\ref{fig:overall} illustrates the overall architecture of our model. We consider a configuration where multi-view camera images and radar point clouds are used for object detection.  Radar points and camera images are passed through separate backbone networks to generate feature maps for each sensor. The Radar-Guided BEV Encoder efficiently converts the camera view features to BEV and fuses them with the radar BEV features. Then, initial 3D proposals are obtained from the fused BEV feature maps using the transformer-based detection head. Finally, the Radar Grid Point Refinement module is used to refine the initial results, maximize the utilization of the radar data, and improve object detection accuracy.

\subsection{Radar and Camera Backbones}
Prior to combining radar and camera features, features are extracted using  separate backbone networks for radar and camera modalities. Multi-scale image features $F_{C}$ are generated from each of multi-view camera images using a ResNet backbone with FPN \cite{FPN}, which are commonly employed in existing camera-based 3D object detection models. For radar data, a PillarNet backbone \cite{pillarnet}  is used to obtain the radar feature map in BEV domain. The preprocessed radar points represented by 3D coordinates $(x, y, z)$, RCS, and Doppler velocity $(v_{x}$, $v_{y})$ are encoded by the 2D sparse convolution and passed through 2D convolution layers to obtain the radar BEV feature map $F_{R}$ $\in \mathbb{R}^{H\times W \times C}$.

\subsection{Radar Guided BEV Encoder}
The Radar Guided BEV Encoder utilizes $F_{R}$ and $F_{C}$ as inputs to generate enhanced BEV features. First, we use $F_{R}$ to create a Radar Guided BEV Query (RGBQ) containing positional information from the radar. Then, we use the RGBQs to transform $F_{C}$ features into enhanced BEV features. Finally, the Radar-Camera Gating (RCG) performs the gated aggregation of enhanced BEV and encoded $F_{R}$ features taking into account the difference in the amount of information between the two BEV features.

\textbf{Radar Guided BEV Query.} In order to achieve feature-level fusion of the radar and camera data, we need to transform $F_{C}$ to a BEV representation. Our baseline model, BEVFormer \cite{BEVFormer}, leverages the BEV query concept and transformer-based structure to perform view transformation. To further boost the accuracy of this view transformation, depth information in radar data can be utilized. Towards this goal, we use the positional information provided by the radar BEV feature map $F_R$ to obtain an enhanced BEV query termed the Radar Guided BEV query $Q^{RG}$ $\in \mathbb{R}^{H\times W \times C}$. This Radar Guided BEV query guides the camera-view to BEV feature transformation. Specifically, the Radar Guided BEV query $Q^{RG}$ is generated by concatenating $F_{R}$ with the learnable BEV queries $Q$ $\in \mathbb{R}^{H \times W \times C}$ and passing it through a deformable self-attention (DeformAttn) module \cite{DeformableDETR} as
\begin{align}
    Q_{p}^{RG} = \sum_{V \in \{Q, F_{R}\}} \text{DeformAttn}(Q_{p}, p, V),
\end{align}
where $Q_{p}^{RG}$ and $Q_{p}$ indicate the queries located at BEV plane pixel $p = (x, y)$. Then, $Q_{p}^{RG}$ is used to decode the camera features through spatial-cross attention (SCA) block \cite{BEVFormer}, generating the refined camera BEV features $B_{C}$ as
\begin{align}\label{SCA}
    B_{C} &= \text{SCA}(Q_{p}^{RG}, F_{C}),
\end{align}
where SCA block is an operation that projects $Q_{p}^{RG}$ to image features and then performs deformable cross-attention \cite{DeformableDETR}.

\textbf{Radar-Camera Gating.} Refined BEV feature map $B_{C}$ and encoded $F_{R}$ are fused by their weighted combination
\begin{align}
    \label{encoded FR}
    F'_{R} &= \text{MLP}_{0}(F_{R}),  \\
    B_{RC} =&\{\sigma (\text{Conv}_{C}[B_{C} \oplus F'_{R}]) \odot B_{C}\} \nonumber \\ &\oplus \{\sigma (\text{Conv}_{R}[F'_{R} \oplus B_{C}]) \odot F'_{R}\},
\end{align}
where $B_{RC}$ denotes the fused BEV feature map, $F'_{R}$ denotes encoded radar BEV feature map, $\text{MLP}(\cdot)$ denotes the channel-wise MLPs, $\sigma(\cdot)$ denotes a sigmoid function, $\odot$ and $\oplus$ denote the element-wise multiplication and summation, respectively. $\text{Conv}_{C}$ and $\text{Conv}_{R}$ are convolutional layers for camera and radar, respectively. The gating operation adaptively weights the $F_{R}$ and $B_{C}$ taking into account information from both sides. This gating operation is implemented using a sigmoid function followed by convolutional layers. 

Next, the fused BEV feature map $B_{RC}$ is passed through normalization and feed-forward network similar to our baseline models, and the final BEV feature map is produced by repeating the BEV encoder section $L$ times. When using the Radar Guided BEV Encoder, an enhanced feature map can be obtained based on radar positional information compared to the existing camera-only method. As seen in the activation maps in Figure 3, the feature map produced by our RCM-Fusion can resolve the region of the objects better than the baseline, BEVFormer-S \cite{BEVFormer}, which does not use radar information for BEV transformation. 

\subsection{Radar Grid Point Refinement}
In this section, we propose a novel instance-level fusion based on the Radar Grid Point Refinement module. The Proposal-aware Radar Attention (PRA) takes 3D proposals and their associated radar points as input and utilizes MLP-based attention layers to determine the importance of each radar point. Then, Radar Grid Point Pooling (RGPP) considers the characteristics and distribution of radar points to sample the virtual grid points, and integrates the radar point features and multi-scale image features into the grid points for refinement. Finally, the refined features and the initial proposal features are combined to produce the final output.

\begin{figure}
    \centering
        \includegraphics[width=0.99\linewidth]{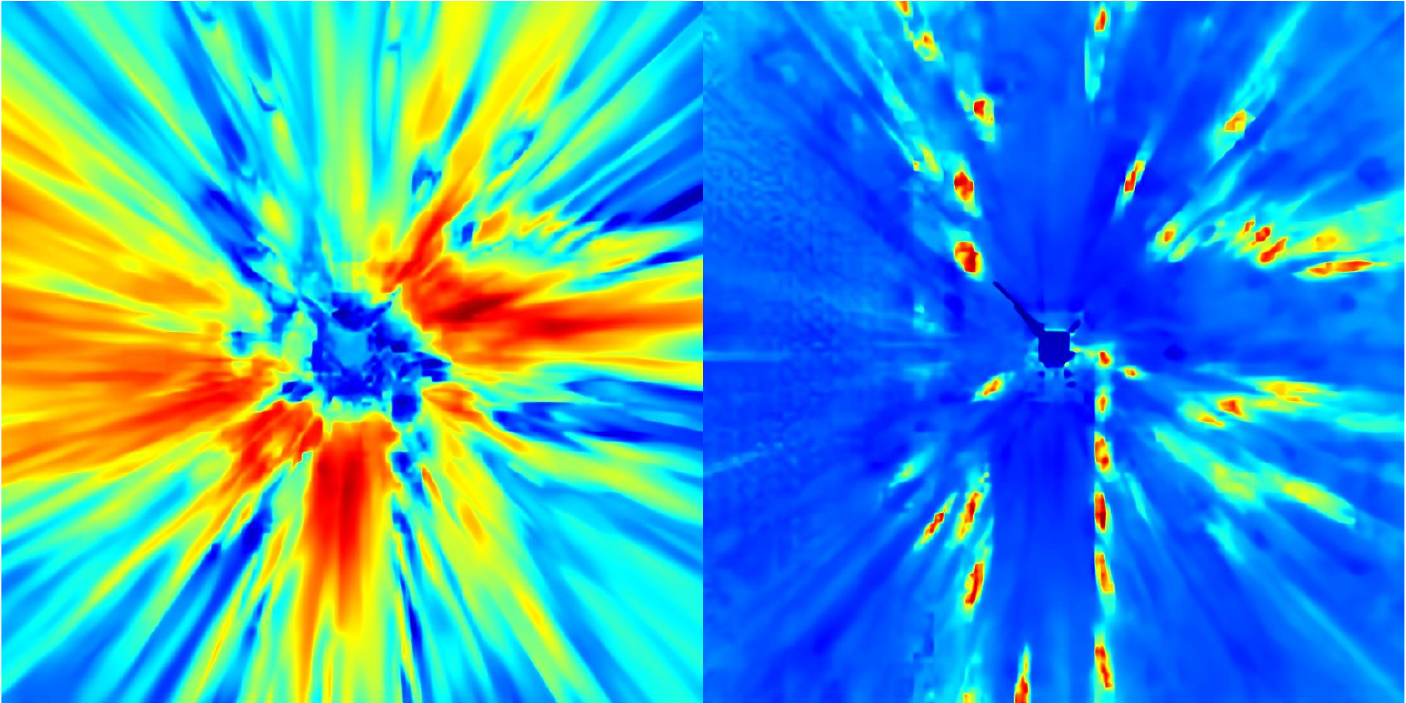}
        \caption{\textbf{Visualization of the BEV feature map:} Left: BEV feature map from BEVFormer-S \cite{BEVFormer}, a single-frame version of BEVFormer. Right: BEV feature map obtained from Radar Guided BEV Encoder. Compared to the left figure, the right figure shows that the positional information provided by the radar BEV feature map allows the features to better localize the regions where objects exist.}
    \label{fig:feat_comp}
\end{figure}

\textbf{Proposal-aware Radar Attention.} First, we associate the radar points with 3D proposals using  Soft Polar Association (SPA) method \cite{craft}. The SPA method transforms 3D proposals and radar points into polar coordinates, and maps radar points within a fixed azimuth and radial range to the 3D proposal. As the range for associating radar points expands, more radar points can contribute to the refinement. However, some of these points might not be relevant to the 3D proposal. PRA module can be used to solve this problem. Let $b$ denote one of the $N$ proposals found by the first stage detection, where $b = (\mathbf{c}, w, l, h, \theta, \mathbf{v}_\mathit{pred})$ consists of 3D proposal center location $\mathbf{c}$, 3D dimension $(w, l, h)$, orientation $\theta$ and velocity $\mathbf{v}_\mathit{pred}$. 
The $K$ radar points associated with $b$ are denoted by $\{r_{k}\}_{k=1}^{K}$, and the 3D position of the $k$th point $r_{k}$ is $\mathbf{u}_{k}$ $\in$ $\mathbb{R}^{3}$. We encode the radar point features and relative positions between $\mathbf{c}$ and $\mathbf{{u}}_{k}$ to calculate the importance weight $s_{k}$ for the $k$th point and obtain the attended radar point features $a_{k}$ as
\begin{align}\label{score_vector2}
    s_{k} & =  \text{MLP}_{2}\left(\left[\text{MLP}_{1}(r_{k}) ; \delta(\mathbf{c} -\mathbf{u}_{k})\right]\right)\\
    \label{attn_feat2}
    a_{k} &= \text{softmax}(\{s_{k}\}_{k=1}^{K})_{k} \odot \text{MLP}_{3}(r_{k}),\\
    & \begin{aligned}
        \notag \hspace{1cm} \text{where } & \forall k \in \{1, \cdots, K\} 
    \end{aligned}
\end{align}
where $\delta(\cdot)$ denotes the positional encoding and $\mathit{softmax}(\cdot)$ denotes the Softmax function.

\begin{figure}
    \centering
        \includegraphics[width=0.99\linewidth]{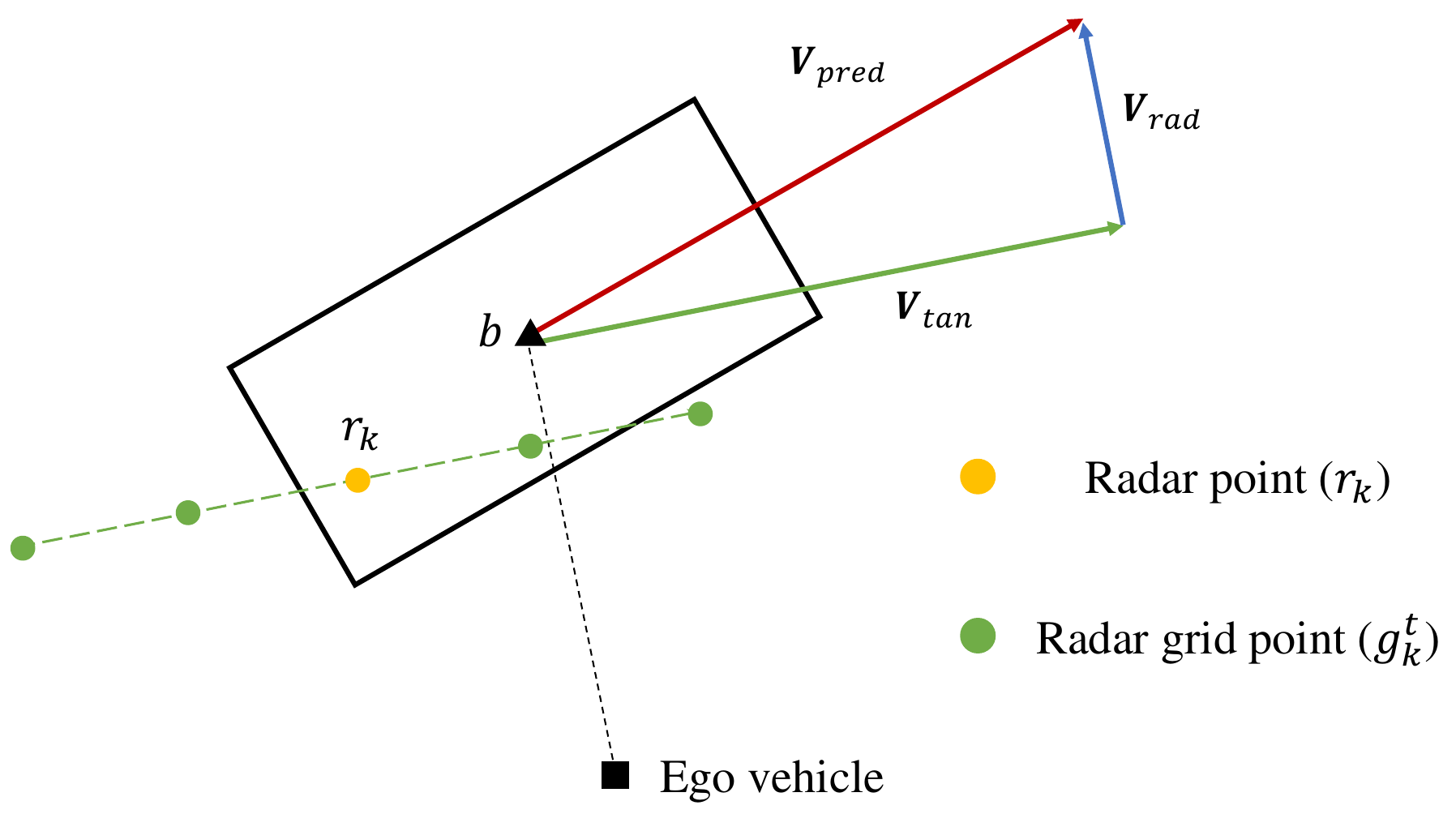}
        \caption{\textbf{Illustration of novel adaptive grid point sampling:} For each radar point associated with the proposal, $T$ grid points are generated in a tangential direction.}
        \label{fig:long}
    \label{fig:onecol}
\end{figure}
\newcolumntype{C}{>{\centering\arraybackslash}p{2.3em}}
\newcolumntype{'}{!{\vrule width 0.1pt}}
\renewcommand{\arraystretch}{1.0}

\begin{table*}[t]\caption{Performance comparisons with single frame 3D object detector on the nuScenes test set. `C', and `R' represent camera, and radar, respectively. * are trained with external data.}
\begin{center}

\begin{adjustbox}{width=0.95\linewidth}

{
\fontsize{4pt}{5pt}\selectfont
\begin{tabular}{c ' c ' c ' C  C '  C  C  C  C  C}
\Xhline{2\arrayrulewidth}

Method & Input & Backbone & NDS & mAP & mATE & mASE & mAOE & mAVE & mAAE\\ 
\Xhline{0.1\arrayrulewidth}

BEVDet* \cite{BEVDet} & \multirow{7}{*}{\begin{tabular}[c]{@{}c@{}}C\end{tabular}} & V2-99 & 48.8 & 42.4 & 0.524 & 0.242 & 0.373 & 0.950 & 0.148 \\
PETR \cite{PETR} &  & R101  & 45.5 & 39.1 & 0.647 & 0.251 & 0.433 & 0.933 & 0.143 \\
Graph-DETR3D \cite{Graph-DETR3D} &  & R101  & 47.2 & 41.8 & 0.668 & 0.250 & 0.440 & 0.876 & 0.139 \\
Ego3RT \cite{Ego3RT} &  & R101 & 44.3 & 38.9 & 0.599 & 0.268 & 0.470 & 1.169 & 0.172 \\
BEVFormer-S \cite{BEVFormer} &  & R101 & 46.2 & 40.9 & 0.650 & 0.261 & 0.439 & 0.925 & 0.147 \\
PolarFormer \cite{PolarFormer} &  & R101 & 47.0 & 41.5 & 0.657 & 0.263 & 0.405 & 0.911 & 0.139 \\
M{\scalebox{0.7}{$^2$}BEV \cite{m2bev}} &  & X101 & 47.4 & 42.9 & 0.583 & 0.254 & 0.376 & 1.053 & 0.190 \\ 
\Xhline{0.1\arrayrulewidth}

CenterFusion \cite{Centerfusion} & \multirow{4}{*}{\begin{tabular}[c]{@{}c@{}}R+C\end{tabular}} & DLA34 & 44.9 & 32.6 & 0.631 & 0.261 & 0.516 & 0.614 & 0.115 \\
CRAFT \cite{craft} &  & DLA34 & 52.3 & 41.1 & 0.467 & 0.268 & 0.456 & 0.519 & 0.114 \\
RCM-Fusion-R50 &  & R50 & \textbf{53.5} & \textbf{45.2} & 0.496 & 0.264 & 0.476 & 0.554 & 0.122 \\
RCM-Fusion-R101 &  & R101 & \textbf{58.7} & \textbf{50.6} & 0.465 & 0.254 & 0.384 & 0.438 & 0.121 \\

\Xhline{2\arrayrulewidth}

\end{tabular}
}
\end{adjustbox}
\end{center}
\label{table:sota}
\end{table*}
\renewcommand{\arraystretch}{1}

\renewcommand{\arraystretch}{1.0}

\begin{table}[t]
\caption{Ablation study conducted on nuScenes val set. RGBQ: the Radar Guided BEV Query. RCG: the Radar-Camera Gating. RGPP: the Radar Grid Point Pooling. PRA: the Proposal-aware Radar Attention.}
\label{table:ablation_main}
\begin{center}
\begin{adjustbox}{width=0.45\textwidth}
{
\fontsize{30pt}{40pt}\selectfont
\begin{tabular}{cccc|cccc}
\Xhline{15\arrayrulewidth}
\multicolumn{2}{c|}{Feature Level} & \multicolumn{2}{c|}{Instance Level} & \multicolumn{4}{c}{Performance} \\ \hline
w/ RGBQ & \multicolumn{1}{c|}{w/ RCG} & w/ RGPP & w/ PRA & NDS & mAP & FPS & Params \\ \hline
 &  &  &  & 44.9 & 37.9 & 2.7 & 66.24M \\
\checkmark &  &  &  & 49.4 & 42.5 & 2.5 & 85.44M \\
\checkmark & \checkmark &  &  & 51.6 & 43.2 & 2.4 & 87.41M \\
\checkmark & \checkmark & \checkmark & & 52.3 & 43.7 & 2.3 & 87.60M \\
\checkmark & \checkmark & \checkmark & \checkmark & \textbf{52.9} & \textbf{44.3} & 2.3 & 87.62M\\
\Xhline{15\arrayrulewidth}
\end{tabular}
}
\end{adjustbox}
\end{center}
\end{table}
\renewcommand{\arraystretch}{1.0}

\begin{table}[t]
\caption{Comparison of different grid point sampling methods. w/o Grid Points refers to a method of substituting radar points with grid points. Fixed Grid Points refers to a method that generates a fixed number of grid points \cite{pvrcnn}.}
\label{table:ablation_gp}
\begin{center}
\begin{adjustbox}{width=0.45\textwidth}
{
\fontsize{3pt}{4pt}\selectfont
\begin{tabular}{c 'c c}
\Xhline{1.5\arrayrulewidth} 
Method & NDS & mAP \\ 
\Xhline{0.1\arrayrulewidth}
w/o Grid Points      & 52.4 & 43.8 \\
Fixed Grid Points  & 52.5 & 44.0 \\
Adaptive Grid Points & \textbf{52.9} & \textbf{44.3} \\
\Xhline{1.5\arrayrulewidth}
\end{tabular}
}
\end{adjustbox}
\end{center}
\end{table}

\textbf{Radar Grid Point Pooling.} The positioning and quantity of grid points have the most significant impact on the efficacy of the grid point-based refinement module \cite{voxelrcnn, PyramidRCNN, pvrcnn}. In particular, radar points have a very low point density and a high positional error compared to LiDAR, which makes the grid point sampling more critical. Therefore, we propose the RGPP module, an adaptive grid point pooling technique that takes into account the characteristics of radar points. As shown in Fig.~\ref{fig:onecol}, the velocity vector $\mathbf{v}_\mathit{pred}$ of 3D proposals, can be decomposed as the tangential velocity $\mathbf{v}_\mathit{tan}$ and the radial velocity $\mathbf{v}_\mathit{rad}$. For the $k^{th}$ radar point $r_k$, $T$ grid points $\{g_{k}^{t}\}_{t=0}^{T-1}$ are generated around the radar point's position $\mathbf{u}_k$ according to
\begin{align}\label{gen_range}
     \gamma &= \left\{ \begin{array}{ccc} \rho_\mathit{min}, &   \quad |\mathbf{v}_\mathit{tan}| \le \rho_\mathit{min} \\ 
   |\mathbf{v}_\mathit{tan}|, &  \rho_\mathit{min} < |\mathbf{v}_\mathit{tan}| < \rho_\mathit{max} \\
   \rho_\mathit{max}, & |\mathbf{v}_\mathit{tan}| \ge \rho_\mathit{max} \end{array} \right.  \\
\label{gen_grid}
    g_{k}^{t} &= \gamma \cdot \left({\frac{t}{T-1}} - {\frac{1}{2}}\right) \cdot {\frac{\mathbf{v}_\mathit{tan}}{|\mathbf{v}_\mathit{tan}|}} + \mathbf{u}_{k}, t = 0, ..., T-1.
\end{align}
As shown in Fig.~\ref{fig:onecol}, we create the grid points in the direction of $\mathbf{v}_\mathit{tan}$. This arrangement is because radar points tend to be noisier in the tangential direction, so the grid points need to be located in this direction to accommodate the best radar point features.
The spacing between the adjacent grid points is determined by $\gamma$, which is proportional to the magnitude of $\mathbf{v}_\mathit{tan}$. Since $\mathbf{v}_\mathit{tan}$ have a predicted value, it may not be accurate, and grid points may not be generated correctly. Therefore, we set the upper bound $\rho_\mathit{max}$ and the lower bound $\rho_\mathit{min}$ for the range of $\gamma$. For each 3D proposal, $\mathit{KT}$ grid points are generated from $K$ radar points, and then Farthest Point Sampling (FPS) is utilized to select $M$ grid points $\{g_{m}\}_{m=1}^{M}$ to maintain a fixed number of grid points.

Next, we use the grid points $\{g_{m}\}_{m=1}^{M}$ as a basis for extracting radar point features and image features and generating the fused grid features. First, the set abstraction (SetAbs) \cite{pvrcnn} encodes the radar points around each grid point $g_{m}$ to produce the radar point features $F_{m}^{pts}$
\begin{align}\label{SetAbs}
    F_{m}^{pts} &= \text{SetAbs}(\{a_{k}\}_{k=1}^{K}, \{u_{k}\}_{k=1}^{K} ,g_{m}).
\end{align}
In parallel, the grid points are projected onto $F_{C}$ in the camera view and then the camera features $F_{m}^{img}$ are obtained through a bilinear interpolation 
 \begin{align}
   \label{bilinear}
    F_{m}^{img} &= \text{Bilinear}(F_{C}, \text{proj}(g_{m})), 
\end{align}
where $\text{proj}(\cdot)$ denotes the projection matrix to the camera view.
Finally, the proposal features $F_{m}^{obj}$ are obtained by fusing $F_{m}^{pts}$ and $F_{m}^{img}$
\begin{align}
    \label{obj_feat}
    F_{m}^{obj} &= \text{maxpool}(F_{m}^{pts} \oplus F_{m}^{img}),
\end{align}
and these features are fused with the initial proposal features to perform 3D proposal refinement.

\renewcommand{\arraystretch}{1.0}

\begin{table}[ht]
    \centering
    \caption{Performance analysis on nuScenes val set.}
    \label{table:hyper}
    \begin{tabular}{c|ccc}
    \Xhline{4\arrayrulewidth}
    \multirow{2}{*}{Metric} & \multicolumn{3}{c}{\# of Encoder Layers}\\ \cline{2-4}
    & 1 & 2 & 3 \\ \hline
    NDS    & 51.2    & \textbf{52.9}    & 52.7 \\
    mAP    & 42.3    & \textbf{44.3}    & 44.1 \\
    \Xhline{4\arrayrulewidth}
    \end{tabular}

    \vspace{0.5em}
    \textbf{(a) \# of Encoder Layers $L$}
    \vspace{0.5em}
    
    \begin{minipage}[t]{0.45\columnwidth}
        \centering
        \begin{tabular}{m{2.5em}|c@{\hspace{0.25cm}}c@{\hspace{0.25cm}}c@{\hspace{0.24cm}}c@{\hspace{0.1cm}}}
        \Xhline{4\arrayrulewidth}
           \multirow{2}{*}{Metric} & \multicolumn{4}{c}{\# of Grid Point} \\ \cline{2-5}
            & 3    & 5    & 7    & 9         \\ \hline
            NDS    & 52.6    & 52.7    & \textbf{52.9}    & 52.9 \\
            mAP    & 44.0    & 44.2    & \textbf{44.3}    & 44.2 \\ 
           \Xhline{4\arrayrulewidth}
        \end{tabular}\\[0.5em]
        \textbf{(b) \#  of Grid Points Per Radar Point, $T$ }
    \end{minipage}
    \hfill
    \begin{minipage}[t]{0.5\columnwidth}
        \centering
        \begin{tabular}{m{2.5em}|c@{\hspace{0.25cm}}c@{\hspace{0.25cm}}c@{\hspace{0.24cm}}c@{\hspace{0.1cm}}}
        \Xhline{4\arrayrulewidth}
           \multirow{2}{*}{Metric} & \multicolumn{4}{c}{\# of Grid Point FPS} \\ \cline{2-5}
            & 16    & 32    & 64    & 128         \\ \hline
            NDS & 52.4 & 52.5 & \textbf{52.9} & 52.4 \\
            mAP & 44.1 & 44.3 & \textbf{44.3} & 44.0 \\ 
        \Xhline{4\arrayrulewidth}
        \end{tabular}\\[0.5em]
        \textbf{(c) \# of Selected Grid Points, $M$}
    \end{minipage}
\end{table}

\section{EXPERIMENTS}

\subsection{Datasets}
We evaluated our model on the nuScenes \cite{nuscenes} dataset, a benchmark for autonomous driving. The nuScenes dataset consists of 700, 150, and 150 scenes for \textit{train}, \textit{val}, and \textit{test}, respectively. The nuScenes dataset  evaluates 3D object detection performance using mean Average Precision (mAP) and the nuScenes Detection score (NDS). The mAP is calculated from the prediction and ground truth's BEV center distance, averaged over 0.5, 1, 2, and 4 meters thresholds. The nuScenes Detection Score (NDS) is calculated by combining mAP with five True Positive (TP) metrics: translation, scale, orientation, velocity, and attribute errors.

\subsection{Implementation Details}
As a baseline model, we used a static version of BEVFormer \cite{BEVFormer}, BEVFormer-S, that only utilized single-frame data. We used the pre-trained weights of FCOS3D \cite{FCOS3D} as a camera backbone. We refer to our method as RCM-Fusion-R101 when ResNet-101 is used.  We also call RCM-Fusion-R50 when ResNet-50 is used. We used the image size of 900 $\times$ 1600 for RCM-Fusion-R101  and the size of 450 $\times$ 800 for RCM-Fusion-50. For radar, we accumulate six previous radar sweeps.
We did not use pre-trained weights for the radar backbone and trained the entire model with random initialization in an end-to-end fashion. We applied the Image Data Augmentation and BEV Data Augmentation techniques in BEVDet \cite{BEVDet}. We used the AdamW \cite{AdamW} optimizer with a learning rate of 2e-4 and trained 24 epochs, employing the class balancing strategy CBGS proposed in \cite{cbgs}. All ablation studies used 1/7 of the nuScenes \textit{train} and applied the CBGS technique for RCM-Fusion-R101.

\begin{figure*}[t]
	\centering
        \centerline{\includegraphics[width=1.0\textwidth]{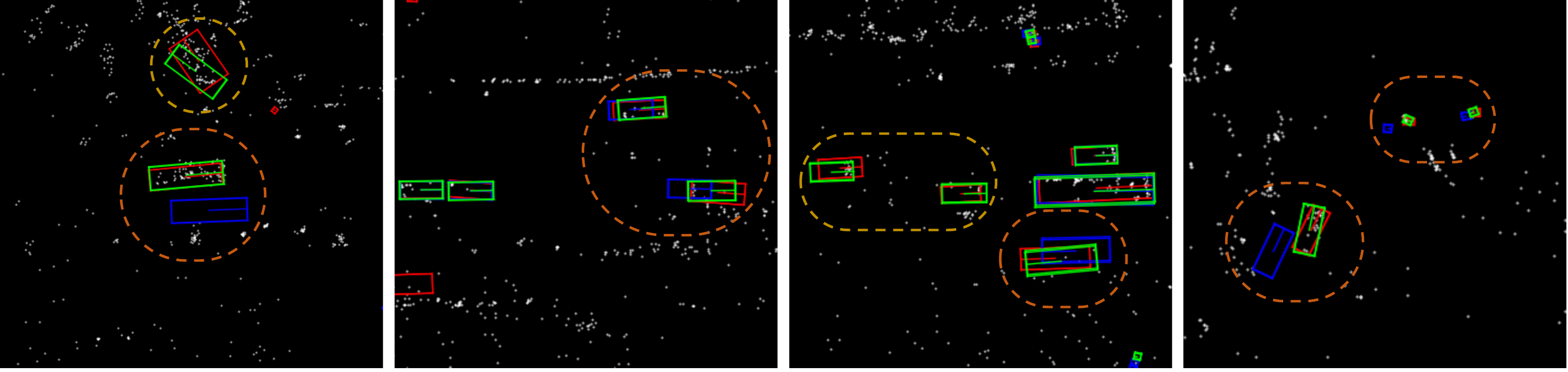}}
    	\caption {\textbf{Qualitative results:} We visualized the predictions of RCM-Fusion (green boxes) and camera-only baseline model, BEVFormer-S (blue boxes), alongside the GT (red boxes) on a bird's eye view. The figures demonstrate that the RCM-Fusion produces more accurate results (in the orange regions), and it can detect the objects the camera-only baseline fails to identify (in the yellow regions).}
	\label{visualize}
\end{figure*}

\subsection{Results on the nuScenes Dataset}
Table~\ref{table:sota} presents the performance of our RCM-Fusion on the nuScenes \textit{test} set. We include the performance of the existing camera-based methods and radar-camera fusion methods. For a fair comparison, we focus solely on single frame-based methods. Thus, CRN \cite{crn} was excluded because its single frame performance was not provided in the original paper. Compared with our baseline model, BEVFormer-S \cite{BEVFormer}, RCM-Fusion-R101 achieves 12.5\% and 9.7\% performance gains in NDS and mAP, respectively. Note that RCM-Fusion-R101 outperforms all other camera-based and radar-camera fusion methods by significant margins. Despite the ResNet50 backbone being considered inferior to the DLA34 backbone, RCM-Fusion-R50 also outperforms CRAFT \cite{craft} by 1.2\% in NDS and 4.1\% in mAP.

\subsection{Ablation Studies on the nuScenes Val Set}
\textbf{Component Analysis.}
Table~\ref{table:ablation_main} presents the ablation study that identifies the contributions of Radar Guided BEV Query (RGBQ), Radar-Camera Gating (RCG), Radar Grid Point Pooling (RGPP) and Proposal-aware Radar Attention (PRA) to the overall performance. Utilizing radar pillar features to refine BEV queries, RGBQ improves the performance of the baseline by 4.6\% in mAP and 4.5\% in NDS. RCG that adaptively fuses multi-modal BEV features offers additional performance gains of 0.7\% in mAP and 2.2\% in NDS. In the instance-level fusion, RGPP, which fuses the multi-modal features based on the adaptive grid points, further improves the performance by 0.5\% in mAP and 0.7\% in NDS. Finally, adding the PRA module leads to a 0.6\% performance gain in both mAP and NDS.  Note that the instance-level fusion results in only a slight increase in both frames per second (FPS) and the number of parameters while it achieves an 1.3\% gain in NDS and 1.0\% gain in mAP. These results demonstrate the effectiveness of our instance-level fusion.

\textbf{Qualitative Results.}
In Fig.~\ref{visualize}, the detection results from RCM-Fusion are visualized for various scenarios. By utilizing the range information provided by the radar, RCM-Fusion can localize the objects more accurately than the camera-only baseline model, BEVFormer-S. Moreover, our model can detect some objects that the camera-only baseline model fails to identify.

\textbf{Effects of Radar Grid Point Sampling.}
We investigate the effectiveness of the proposed grid point generation method. Table~\ref{table:ablation_gp} compares the proposed method with other grid generation methods. The w/o Grid Points (w/o GP) is a method that does not generate grid points and directly uses the radar points associated with the proposal for refinement. The Fixed Grid Point method is a method that generates a fixed number of grid points regularly distributed within a proposal box, similar to \cite{pvrcnn}. While the Fixed Grid Point method was effective for LiDAR data, its performance is not satisfactory with radar data because of its sparse distribution and increased errors along the horizontal axis. We note that the Fixed Grid Point method barely improves the w/o GP method. However, by employing Adaptive Grid Points that adjust both direction and spacing based on the distribution of radar points, our grid point sampling method achieved better performance over the baselines.

\textbf{Performance versus Hyper-parameters.}
Table~\ref{table:hyper} presents the performance of RCM-Fusion in relation to several hyper-parameters. Specifically, Table~\ref{table:hyper} (a) investigates the performance as a function of the number of layers in the Radar Guided BEV encoder. Our experimental results indicate that a performance peak is achieved when the number of layers is set to two. Table~\ref{table:hyper} (b) presents the performance versus the number of grid points $T$ generated per each radar point. We found that the best value for $T$ is 7.  Table~\ref{table:hyper} (c) shows the best value for the number of grid points $M$ used in the FPS algorithm is 64.

\section{CONCLUSIONS}
In this paper, we proposed a novel radar-camera 3D detector, RCM-Fusion. RCM-Fusion fused the radar and camera data at both the feature-level and instance-level for enhanced 3D object detection. We demonstrated that the Radar Guided BEV Encoder could achieve effective transformation of camera features into BEV, leveraging the radar features. Moreover, the Radar Grid Point Refinement proposed a novel grid pooling technique, which fused both radar and camera features based on the adaptive grid points to refine the initial 3D proposals. For future work, we plan to expand our approach towards designing a time-efficient multi-frame radar-camera fusion method.

\section{ACKNOWLEDGMENT}
This work was supported by Institute for Information \& communications Technology Promotion(IITP) and the National Research Foundation of Korea (NRF) grant funded by the Korea government (MSIT) (No.2021-0-01314,Development of driving environment data stitching technology to provide data on shaded areas for autonomous vehicles and 2020R1A2C2012146)

\bibliographystyle{plain}
\bibliography{Reference}

\end{document}